\documentclass{article}
\usepackage{spconf,amsmath,graphicx}
\usepackage[ruled,vlined]{algorithm2e}

\usepackage[style=ieee,maxbibnames=4]{biblatex}
\usepackage{enumitem}
\setlist{nosep, leftmargin=14pt}

\usepackage{mwe} % to get dummy images

\title{DermAI: Clinical dermatology acquisition through quality-driven image collection for AI classification in mobile}
\name{
\parbox{\linewidth}{\centering
Thales Bezerra$^{\star}$ \qquad Emanoel Thyago$^{\star}$ \qquad Kelvin Cunha$^{\star}$ \qquad Rodrigo Abreu$^{\star}$ \qquad Fábio Papais$^{\star}$ \qquad Francisco Mauro$^{\star}$ \qquad Natália Lopes$^{\star}$ \qquad Érico Medeiros$^{\star}$ \qquad Jéssica Guido$^{\dagger}$ \qquad Shirley Cruz$^{\dagger}$ \qquad Paulo Borba$^{\star}$ \qquad Tsang Ing Ren$^{\star}$
}}

\address{$^{\star}$ Centro de Informática, Universidade Federal de Pernambuco, Brazil \\
     $^{\dagger}$ Hospital das Clínicas, Universidade Federal de Pernambuco, Brazil}
     
\begin{document}
%\ninept
%
\maketitle
\begin{abstract}

AI-based dermatology adoption remains limited by biased datasets, variable image quality, and limited validation. We introduce DermAI, a lightweight, smartphone-based application that enables real-time capture, annotation, and classification of skin lesions during routine consultations. Unlike prior dermoscopy-focused tools, DermAI performs on-device quality checks, and local model adaptation. The DermAI clinical dataset, encompasses a wide range of skin tones, ethinicity and source devices. In preliminary experiments, models trained on public datasets failed to generalize to our samples, while fine-tuning with local data improved performance. These results highlight the importance of standardized, diverse data collection aligned with healthcare needs and oriented to machine learning development. The dataset will be made available for research purposes upon request to the corresponding authors.

\end{abstract}
\section{Introduction}
\label{sec:intro}

The development of AI solutions in healthcare has rapidly increased, particularly in medical image analysis~\cite{hasan2023survey}. In dermatology, such models can support clinicians by providing preliminary lesion assessments~\cite{yan2025multimodal,zeng2025mm}. However, despite the availability of several public datasets~\cite{Daneshjou2022-cd,Pacheco2020-mn,tschandl2018ham10000}, many suffer from limited diversity, lack of acquisition standards, and imaging artifacts, which reduce their clinical applicability~\cite{daneshjou2021lack}. Moreover, while these datasets consider clinical challenges, they rarely address practical constraints required for machine learning pipelines. Consequently, the models trained often fail to generalize to real-world conditions, especially in public healthcare systems serving diverse populations.

To address these limitations, we introduce DermAI, designed for standardized dermatological image acquisition oriented to AI training and metadata collection in clinical environments. The platform enforces consistent capture guidelines and performs real-time quality checks. Designed to support dermatologists and general practitioners, it facilitates early triage and risk assessment while generating clinically meaningful training data. Using the platform, we constructed a new clinical dataset collected during routine consultations at a Brazilian public university hospital, encompassing a wide range of skin tones, ethnicities, devices, and environmental conditions. We benchmarked convolutional neural networks on this dataset and compared their performance to models trained on public datasets; Models trained on public data showed limited generalization under real clinical variability. This result emphasize the need for context-specific samples and demonstrate DermAI as a practical acquisition tool.

\section{Available datasets}

Several image-based dermatology datasets have supported the development of learning-based skin lesion classification models. These datasets are typically categorized into clinical~\cite{Daneshjou2022-cd,Pacheco2020-mn} and dermoscopic~\cite{tschandl2018ham10000}. However, they present critical limitations~\cite{Daneshjou2022-cd,hasan2023survey}, including restricted access, unclear acquisition protocols, and underrepresentation of diverse skin tones and ethnic groups, hindering fairness and model generalization~\cite{wu2021medical}. The ISIC Archive\footnote{\url{https://www.isic-archive.com/}} remains the largest dermoscopic repository with histopathologically confirmed samples, yet is predominantly composed of lighter skin types~\cite{daneshjou2021lack}. Clinical datasets such as DDI~\cite{Daneshjou2022-cd} and PAD-UFES-20~\cite{Pacheco2020-mn} improve diversity but still contain acquisition artifacts, show limited variability across environments and devices, and provide relatively few samples.

\section{DermAI}

\begin{figure*}[htb]
  \centering
  \includegraphics[width=\linewidth]{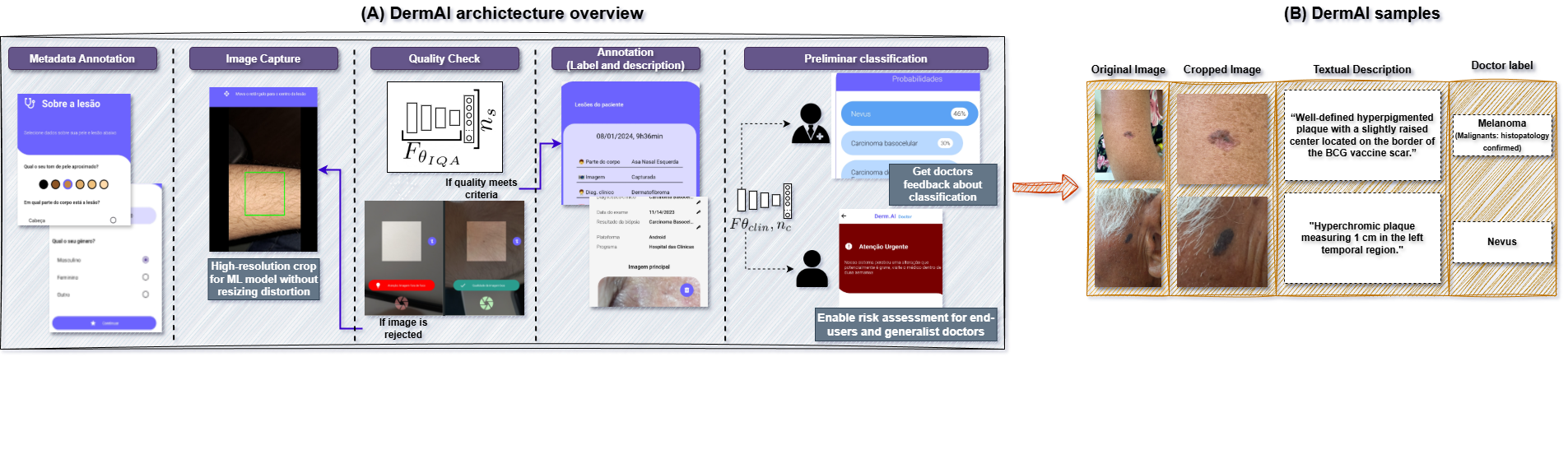}
\caption{An overview on the APP structure. (A) Shows the application main steps: image capture, quality validation, annotation and label review, and preliminar classification. (B) Illustrate the resulting samples,  complete and cropped image, textual image description, and preliminar label.}
\label{figure:app_overview}
\end{figure*}

\textbf{DermAI} is a scalable solution designed to dermatological workflows. It standardizes image acquisition, particularly in low-resource settings where non-specialist users rely on standard mobile devices. Thus, DermAI enables reproducible and controlled image capture suitable for machine learning training under diverse clinical conditions.

\subsection{Acquisition, quality validation and label review}

We collaborated with board-certified dermatologists, residents, and medical students to collect clinically relevant data during consultations at the \textit{Hospital das Clínicas, Universidade Federal de Pernambuco}. Each patient received a unique record identifier to link images and metadata. A standardized form was completed for every case, including patient age, Fitzpatrick skin phototype, gender, lesion location, and lesion mask. Lesion images were captured directly using smartphone cameras and paired with short clinical descriptions, producing image-text annotations (Figure~\ref{figure:app_overview}-b). Device attributes (model, operating system, and camera specifications) were also recorded. Medical students accompanied clinicians during consultations but were responsible only for image acquisition. All samples were reviewed and validated by the attending dermatologist, and the preliminary diagnosis reflects their clinical judgment. To address common issues in clinical dermatology imaging (e.g., ink markings, gel artifacts, or glare), the application provides acquisition guidelines. Clinicians avoid pre-marking, center the lesion using on-screen guides, and capture images at a standardized distance of approximately $5$\,cm. Small or out-of-focus lesions are corrected during preprocessing. Environmental conditions remain unconstrained to reflect clinical variability.

Clinical image acquisition is naturally affected by uncontrolled factors, such as ambient lighting, camera jitter, inconsistent distances or angles, and incorrect focus. To mitigate these effects, we preprocess images by cropping a central region and enforcing a square aspect ratio, which reduces background content, emphasizes the lesion, and minimizes distortion. During acquisition, the application shows a preview of the cropped region to help ensure framing consistency. This design aligns with common machine learning pipelines that require square inputs, limiting geometric distortion from the point of capture. Both the original and cropped images are stored. We also developed a learning-based model for image quality assessment. The model, based on a MobileNetV3 backbone ($F_{\theta_{IQA}}$) followed by a linear layer with SiLU activation, dropout ($0.2$), and a sigmoid output, produces $n_s = 4$ binary indicators corresponding to sharpness, blur, exposure (over/under), and compression artifacts. It was trained using images augmented with synthetic distortions, particularly those acquired from smartphones. Rather than correcting the images, it quantifies image quality and prompts the user to recapture in real time, preventing severely degraded samples from entering the dataset. After acquisition, samples that meet quality criteria are reviewed by dermatologists, who mark the lesion center and define a circular region of interest. Each image contains a single lesion; multiple lesions require separate captures to ensure consistent annotation. Clinicians may flag lesions suspected of malignancy, prompting biopsy follow-up and later inclusion of histopathological results. This workflow ensures that malignant-suspect cases undergo systematic verification, improving the reliability of diagnostic labels in the dataset.

\subsection{Dataset}

\begin{figure}[htb]
  \centering
  \includegraphics[width=0.8\columnwidth]{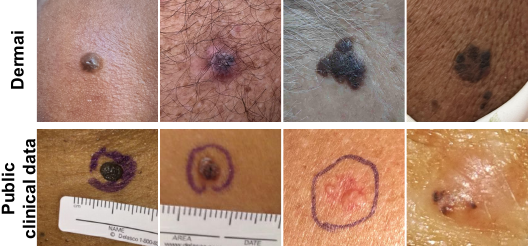}
\caption{Example samples collected with DermAI (top row), illustrating artifact-controlled image acquisition. For comparison, the bottom row shows samples from public datasets, which may include distractors, blur, or overexposure.}
\label{fig:other_samples}
\end{figure}

At the time of this study, data collection with the DermAI has been ongoing for two years. The current dataset comprises $3,401$ images corresponding to $200$ uniquely annotated lesions ($2273$ benign, $608$ malignant, and $520$ pre-malignant), including, melanoma (30), basal cell carcinoma (477), squamous cell carcinoma (142), nevus (568), actinic keratosis (520), benign keratosis (706), and solar lentigo (285). Figure~\ref{fig:other_samples} shows representative examples from our collection and contrasts them with samples from existing datasets, highlighting the reduction of artifacts achieved through our standardized acquisition protocol and diverse skin tone representation. While we aim to expand data collection through the app, the dataset is expected to grow over time. 

\subsection{Preliminar classification and medical feedback}

In the final step, a classification model provides a preliminary suggestion. The model is based on CNNs, following prior work~\cite{Pacheco2020-mn,Daneshjou2022-cd}. Several backbones were evaluated, and the best-performing ones were trained and converted for local on-device inference. The model classifies $n_c$ lesion categories to support clinical decision-making and gather feedback from dermatologists. For general practitioners or non-specialist users, the binary risk assessment (benign vs. potentially malignant) is provided. The classification architecture consists of the CNN backbone followed by average pooling and a fully connected layer trained with categorical cross-entropy. Predictions are shown to clinicians, who can confirm, disagree, or indicate uncertainty, and enter their own diagnostic hypothesis. This feedback enables performance monitoring under real clinical use. All lesion annotations and metadata are retained, allowing later updates with biopsy confirmation; lesions flagged as malignant-suspect are prioritized for biopsy follow-up. A supervisory interface enables board-certified dermatologists review annotation consistency.

To further improve robustness, we evaluate an ensemble inference strategy. Let $\{f_1, f_2, \ldots, f_K\}$ denote $K$ trained classifiers, each producing a probability vector $p_k \in \mathbf{R}^{n_c}$. We consider two schemes: \textbf{(i) Majority Vote} (Eq.~\ref{eq:major}), which emphasizes consensus among models and reduces variance due to classifier-specific biases; \textbf{(ii) Learned Fusion}: training a small MLP $g_\phi(\cdot)$ that receives the concatenated model outputs (Eq.~\ref{eq:feat}), allowing the ensemble to learn correlations between classifier confidence patterns. The MLP consists of two fully connected layers with ReLU activation and dropout, trained using categorical cross-entropy over validation folds. These ensemble strategies improve predictive stability and sensitivity, particularly for malignant-suspect cases, by leveraging complementary decision patterns among architectures.

\begin{equation}
\label{eq:major}
\hat{y} = \arg\max_{c} \sum_{k=1}^{K} \mathbf{1}\big(\arg\max(p_k) = c\big),
\end{equation}

\begin{equation}
\label{eq:feat}
\hat{y} = \arg\max g_\phi([p_1 \Vert p_2 \Vert \cdots \Vert p_K]),
\end{equation}

\section{Experiments and discussions}

\begin{table*}[!ht]
\centering
\caption{Cross-dataset classification performance across architectures evaluated on accuracy (ACC), recall (R), precision (P), and F1-score (F1). The first eight models are trained on PAD-UFES-20 (PAD), and the next eight on DermAI. All models are evaluated on the three target datasets. The final four rows report the mean performance across architectures trained on: (i) PAD-UFES-20, (ii) PAD-UFES-20 (filtered), (iii) DermAI, and (iv) the combination of PAD-UFES-20 (filtered) + DermAI.}
\label{tab:models_results}
\resizebox{\textwidth}{!}{\begin{tabular}{c|cccc|cccc|cccc}
 & \multicolumn{4}{c|}{PAD-UFES-20~\cite{Pacheco2020-mn}} & \multicolumn{4}{c|}{DDI~\cite{Daneshjou2022-cd}} & \multicolumn{4}{c}{DermAI} \\ \hline
                                      & ACC    & R      & P      & F1     & ACC    & R      & P      & F1     & ACC    & R      & P      & F1 \\ \hline
DenseNet (PAD)                        & 0.7533 & 0.6198 & 0.6343 & 0.6280 & \textbf{0.1987} & \textbf{0.1949} & \textbf{0.1270} & \textbf{0.1537} & 0.4648 & 0.5583 & 0.5745 & 0.5531\\
ConvNeXt (PAD)                        & 0.7503 & 0.6523 & 0.7144 & 0.6718 & 0.1862 & 0.1768 & 0.1518 & 0.1180 & 0.4794 & \textbf{0.5719} & \textbf{0.5906} & \textbf{0.5756} \\
MobileNet v3 (PAD)                    & 0.7050 & 0.5383 & 0.6824 & 0.6018 & 0.1829 & 0.1408 & 0.0568 & 0.0644 & 0.4361 & 0.4366 & 0.4436 & 0.4300 \\
EfficientNet v2 (PAD)                 & 0.7554 & 0.6523 & 0.7144 & 0.6718 & 0.1768 & 0.1518 & 0.1060 & 0.1180 & 0.4897 & 0.4824 & 0.4821 & 0.4968 \\
Resnet (PAD)                          & 0.7073 & 0.7072 & 0.6809 & 0.6883 & 0.1280 & 0.1156 & 0.0438 & 0.0590 & \textbf{0.5568} & 0.4824 & 0.4968 & 0.4821 \\ 
Ensemble-major (PAD)                  & 0.7692 & 0.7400 & 0.7925 & 0.7653 & 0.1737 & 0.1105 & 0.0383 & 0.0517 & 0.5274 & 0.4840 & 0.5511 & 0.5153 \\
Ensemble-fusion (PAD)                 & \textbf{0.7795} & \textbf{0.7800} & \textbf{0.8000} & \textbf{0.7898} & 0.1253 & 0.1167 & 0.0842 & 0.0978 & 0.5283 & 0.5466 & 0.5161 & 0.5309 \\ \hline
DenseNet (DermAI)                     & 0.7317 & 0.7386 & 0.7240 & 0.7224 & 0.3467 & 0.3295 & 0.2933 & 0.3103 & 0.8209 & 0.8176 & 0.8209 & 0.8192 \\
ConvNeXt (DermAI)                     & 0.7526 & 0.7317 & 0.7201 & 0.7205 & 0.3953 & 0.3949 & 0.3987 & 0.3867 & 0.8102 & 0.8092 & 0.8102 & 0.8096 \\
MobileNet v3 (DermAI)                 & 0.6317 & 0.6317 & 0.6201 & 0.6205 & 0.2927 & 0.1899 & 0.2077 & 0.1984 & 0.7834 & 0.7788 & 0.7834 & 0.7765 \\
EfficientNet v2 (DermAI)              & 0.6247 & 0.6247 & 0.6233 & 0.6206 & 0.3943 & 0.3383 & 0.3832 & 0.3593 & 0.7567 & 0.7497 & 0.7567 & 0.7531 \\
Resnet (DermAI)                       & 0.6993 & 0.6933 & 0.6835 & 0.6687 & \textbf{0.5050} & 0.3824 & 0.4383 & 0.4008 & 0.7834 & 0.7805 & 0.7834 & 0.7819 \\ 
Ensemble-major (DermAI)               & 0.7902 & 0.7939 & 0.7902 & 0.7910 & 0.4807 & 0.3156 & 0.2949 & 0.3048 & 0.8234 & 0.8234 & 0.8338 & 0.8232 \\
Ensemble-fusion (DermAI)              & \textbf{0.8035} & \textbf{0.8035} & \textbf{0.8033} & \textbf{0.8033} & 0.4903 & \textbf{0.4880} & \textbf{0.4999} & \textbf{0.4938} & \textbf{0.8467} & \underline{\textbf{0.8583}} & \textbf{0.8560} & \textbf{0.8571} \\ \hline
AVG-all models (PAD)                  & 0.7457 & 0.6700 & 0.7170 & 0.6881 & 0.1673 & 0.1439 & 0.0873 & 0.0941 & 0.4975 & 0.5088 & 0.5221 & 0.5119 \\
AVG-all models (PAD\_filter)          & 0.7817 & 0.7812 & 0.7817 & 0.7850 & 0.1583 & 0.0925 & 0.0225 & 0.03601 & 0.5196 & 0.5291 & 0.5278 & 0.5217 \\
AVG-all models (DermAI)               & 0.7184 & 0.7168 & 0.7092 & 0.7067 & 0.4150 & 0.3483 & 0.3594 & 0.3505 & 0.8035 & 0.8025 & 0.8063 & 0.8088 \\
AVG-all models (PAD\_filter + DermAI) & \underline{\textbf{0.8505}} & \underline{\textbf{0.8462}} & \underline{\textbf{0.8803}} & \underline{\textbf{0.8607}} & 0.5698 & 0.4693 & 0.5793 & 0.5185 & \underline{\textbf{0.8511}} & \textbf{0.8420} & \underline{\textbf{0.8754}} & \underline{\textbf{0.8583}} \\
AVG-all models (PAD + DermAI + DDI)   & 0.7971 & 0.7971 & 0.8158 & 0.7961 & \underline{\textbf{0.7778}} & \underline{\textbf{0.7263}} & \underline{\textbf{0.7311}} & \underline{\textbf{0.7286}} & 0.7956 & 0.8089 & 0.7956 & 0.7952
\end{tabular}}
\end{table*}

%\subsection{Assessing classifiers on public datasets}

We trained multiple CNN architectures (Table~\ref{tab:models_results}) on the PAD-UFES-20 (PAD) dataset~\cite{Pacheco2020-mn}, replicating the original setup while adding recent backbones. Since PAD does not define official splits, we established fixed training and test partitions (80/10/10). We then evaluated the resulting models on DDI~\cite{Daneshjou2022-cd} and DermAI to assess out-of-distribution generalization. We focused on lightweight models suitable for on-device execution. The DDI, due to its small size, was used mainly for testing. Evaluation was restricted to the six lesion classes provided in PAD (melanoma, nevus, seborrheic keratosis, actinic keratosis, basal cell carcinoma, and squamous cell carcinoma), which permits direct analysis between the three datasets. Performance was measured using accuracy (ACC), recall (R), precision (P), and F1-score (F1)~\cite{Pacheco2020-mn}. As shown in Table~\ref{tab:models_results}, PAD-trained models align with previous results but show limited generalization when tested on DDI and DermAI, particularly in recall, which is critical for minimizing missed malignant cases. Although PAD is visually diverse, it contains acquisition artifacts (e.g., ink markings, blur) and limited lesion variety or skin tones, which constrain robust feature learning. Retraining the same architectures using DermAI samples improved cross-dataset performance, reflecting the benefits of controlled and diverse acquisition. While it does not achieve the highest scores in every scenario, the model maintains stable performance on PAD and preserves useful knowledge when evaluated on DDI, the most challenging dataset due to its high noise and variability. We further applied DermAI-based quality filtering to PAD (PAD\_filter) and retrained the models. Despite using fewer samples, the PAD\_filter improved performance across backbones, highlighting the importance of curated image features. Combining PAD\_filter with DermAI yielded the best overall results. Adding DDI samples to training did not improve performance, likely due to their scarcity and noise, reinforcing that increasing data quantity alone is insufficient; quality and curated acquisition are essential for training AI models.

\section{Conclusion}

We demonstrate that clinical algorithms often face significant limitations due to domain shifts, exposing the weaknesses of cross-validation across heterogeneous data sources. Consequently, such models cannot be fully trusted in real-world clinical use. Thus, we introduce DermAI, an acquisition framework that ensures high-quality image capture while remaining accessible for diverse acquisition scenarios. DermAI addresses limited diversity, inconsistent image quality, and lack of standardized acquisition through structured, high-quality data collection across diverse populations oriented to train machine learning models. Designed for continuous use, it enables datasets to grow over time, supporting model refinement and improved generalization. Being model-agnostic and compatible with major ML frameworks, DermAI fosters reproducible research and adaptation, laying the groundwork for AI-assisted dermatological care.

\section{Compliance with ethical standards}
\label{sec:ethics}

The study involving human participants was reviewed and approved by the Research Ethics Commitee of the Federal University of Pernambuco (CEP-UFPE), which operates within the national CEP/CONEP system via the Plataforma Brasil platform (CAAE ID 75991523.3.0000.8807). The research was conducted in accordance with the ethical principles embodied in the Declaration of Helsinki and consistent with the Brazilian National Health Council's Resolution 466/12 (and all applicable local statutory requirements). Written informed consent to participate in this study was obtained from all participants.

\section{Acknowledgments}
\label{sec:acknowledgments}

This project was supported by the Ministry of Science, Technology and Innovation of Brazil, with resources from Law No. 8,248, dated October 23, 1991, under the scope of the PPI-SOFTEX, coordinated by Softex and published under RESIDÊNCIA EM TIC 63 – ROBÓTICA E IA – FASE II, DOU 23076.043130/2025-27.

%\bibliographystyle{IEEEbib}
%\bibliography{references}
\printbibliography

\end{document}